\documentclass[preprint,12pt]{elsarticle}

\usepackage{amssymb}
\usepackage[T1]{fontenc}
\usepackage[utf8]{inputenc}
\usepackage{amssymb}
\usepackage{wasysym}
\usepackage{tikz}
\usepackage{comment}
\usepackage{color}
\usepackage{booktabs}
\usepackage[normalem]{ulem}
\usepackage{listings}     
\usepackage{lstautogobble}  
\usepackage{color}          
\usepackage{zi4}            
\usepackage{subcaption}
\usepackage{url}
\usepackage{listings}
\definecolor{bluekeywords}{rgb}{0.13, 0.13, 1}
\definecolor{greencomments}{rgb}{0, 0.5, 0}
\definecolor{redstrings}{rgb}{0.9, 0, 0}
\definecolor{graynumbers}{rgb}{0.5, 0.5, 0.5}
\usepackage{mdframed}       
\usepackage{multirow}
\usepackage{enumitem}
\setlist{leftmargin=4em}

\usepackage{tabularx}
\newcolumntype{C}{>{\centering\arraybackslash}X}
\newcolumntype{R}{>{\raggedleft\arraybackslash}X}
\newcolumntype{L}{>{\raggedright\arraybackslash}X}

\definecolor{red}{HTML}{FD0100}
\definecolor{orange}{HTML}{F76915}
\definecolor{yellow}{HTML}{EEDE04}
\definecolor{green}{HTML}{A0D636}
\definecolor{teal}{HTML}{2FA236}
\definecolor{blue}{HTML}{333ED4}

\usepackage{makecell}

\journal{Neurocomputing}

\begin{document}

\begin{frontmatter}

\title{\emph{torchosr} --- a \textit{PyTorch} extension package for Open Set Recognition models evaluation in \textit{Python}}

\author{Joanna Komorniczak and Paweł Ksieniewicz}

\address{Department of Systems and Computer Networks,\\ Faculty of Information and Communication Technology,\\Wrocław University of Science and Technology,\\Wybrzeże Wyspiańskiego 27, 50-370 Wrocław, Poland}

\begin{abstract}

The article presents the \textit{torchosr} package -- a \textit{Python} package compatible with \textit{PyTorch} library -- offering tools and methods dedicated to Open Set Recognition in Deep Neural Networks. The package offers two state-of-the-art methods in the field, a set of functions for handling base sets and generation of derived sets for the Open Set Recognition task (where some classes are considered unknown and used only in the testing process) and additional tools to handle datasets and methods. The main goal of the package proposal is to simplify and promote the correct experimental evaluation, where experiments are carried out on a large number of derivative sets with various \textit{Openness} and class-to-category assignments. The authors hope that state-of-the-art methods available in the package will become a source of a correct and open-source implementation of the relevant solutions in the domain.

\end{abstract}


\end{frontmatter}

\section{Motivation and significance}

In recent years, the research community has been paying close attention to the Open Set Recognition (OSR) task \cite{geng2020recent}, which combines the classic closed-set classification and the detection of samples coming from unknown classes \cite{scheirer2012toward}. Methods to solve this task are exceedingly demanded in the face of the growing popularity of deep neural networks, whose distinctive feature is unsupervised feature extraction \cite{lecun2015deep}. Methods recognising in Open Space should be able to detect and handle instances that are not coming from the classes used in the training process. This task is not trivial both in the context of method proposals and evaluation.


Samples of known classes (KKC) are present both in the process of training and testing the model. One of the criteria for evaluating OSR methods is the correct classification within these classes. Unknown class samples (UUC), on the other hand, are used only in the process of testing methods \cite{scheirer2012toward}. The task of the algorithms is to mark these samples as instances of unknown classes.

UUC samples and their distribution information are not available for end-use applications. However, UUC sampling is necessary for experimental analysis and method evaluation purposes \cite{geng2020recent}. Two used UUC sampling protocols are (1) Holdout, where the UUC samples come from the same dataset as the KKC but have not been used to train the model, and (2) Outlier, where the UUC samples come from a different dataset and whose representation has been altered so that the algorithm trained on the set containing KKC would be able to process them.

The first works in the field emphasized the importance of evaluating methods on a wide range of \textit{Openness} problems and a diverse number of test and training sets \cite{scheirer2012toward}. In each \textit{Openness}, it is also worth drawing which classes will be selected as KKC and which as UUC. Such vast possibilities of configuring the sets used to evaluate the methods cause a considerable number of derived sets to be obtained from a single set (or a pair of sets with the Outlier protocol). In each of these sets, in order to stabilize the results, it is worth using cross-validation.

Evaluations of existing solutions are often based on analyzing a dataset described by single \textit{Openness} and single assigning classes to particular (KKC and UUC) categories. Existing packages offer ready-made collections with such a single division \cite{yang2022openood}. In addition, many methods based on thresholding and Weibull distributions emphasize the need to optimize the hyperparameters of the methods for a particular task, even specifying suboptimal values as dependent on \textit{Openness} \cite{scheirer2014probability}. Hence, evaluating methods on multiple problem configurations is a highly desirable feature of model performance analysis. The datasets offered in the \textit{torchosr} package, tools specifying the desired number of set configurations, and dividing data into cross-validation folds can contribute to disseminating proper experimental evaluation and its simplification.


A key element related to the algorithm proposals is comparing new solutions with state-of-the-art methods. There is currently no uniform library aggregating the currently available state-of-the-art methods in the field of Open Set Recognition. Moreover, the methods known from the literature are rarely presented jointly with the complete code in the form of an integral module containing the implementation of the algorithm. Therefore, publications presenting algorithms are the only source of knowledge allowing the proper implementation of the reference methods. When comparing algorithms with other solutions, the key is the correct implementation and hyperparameterization of reference methods, which is foremost possible with an open and commonly available source.

Due to the difficulty of implementing and evaluating reference solutions, some survey articles and papers presenting new solutions compare results with those reported in previously published works \cite{geng2020recent,oza2019c2ae}. It should be emphasized that a critical element of a correct experimental analysis is a homogeneous experimental protocol. Therefore testing methods on different subsets or under different conditions should not allow drawing conclusions about the methods' performance.


The presented package is intended for experimental analysis of OSR methods in an environment compatible with the \textit{PyTorch} library in the \textit{Python} programming language. It is worth emphasizing here that \textit{Python} has become a standard in recent years for proposing and evaluating Machine Learning methods \cite{nguyen2019machine} and a significant pool of solutions in the field published in recent years has been implemented and evaluated in this language \cite{rudd2017extreme,bendale2016towards,shu2017doc,neal2018open,joseph2021towards}.

The \textit{torchosr} package contains two basic Open Set Recognition methods for use in Deep Neural Networks -- \textit{Thresholded Softmax} and \textit{Openmax}~\cite{bendale2016towards} -- as well as tools for experimental analysis. The tools allow dataset division into configurations containing KKC and UUC in the Holdout and Outlier evaluation protocol, considering many \textit{Openness} values and multiple class-selection repetitions. In addition, the tool allows splitting the collections into folds in the cross-validation procedure.

The primary motivation is to present a tool facilitating these methods' experimental evaluation and enabling their replication. The first uniform implementation of the \textit{Thresholded Softmax} and \textit{Openmax} methods, available in the \textit{torchosr} package, has a chance to be a publicly available source of a correct and up-to-date implementation. Implemented solutions, potentionally expanded in the future, will also contribute to facilitating the evaluation of methods and ensure that future solutions compare their performance with the exact implementation of state-of-the-art algorithms.

\section{Software description}

This section contains a description of the individual modules of the presented package and an example experiment presenting its basic functionalities. \textit{Torchosr} extends the functionality of the \textit{PyTorch} library and is based on its functionalities.

\subsection{Software architecture}

\begin{figure}[!htb]
    \centering
    \resizebox{.5\textwidth}{!}{\begin{tikzpicture}[
	scale=1.25,
	entry/.style={text width=4cm, align = left},
	block/.style={rounded corners=.5cm},
	label/.style={rotate=90, white, fill=black,rounded corners=.25cm},
	arr/.style={->, thick}
]

	\draw [block] (1,2) rectangle (10,11);
	\node[label] at (1,6.5) () {\bfseries Data};
    
    \node[entry] at (3,10) () {\emph{DataWrapper}};
    \node[entry] at (3, 9) () {\emph{OutlierDataset}};
    
    \node[entry] at (3, 7.5) () {\emph{configure\_division}};
    \node[entry] at (3, 6.5) () {\emph{configure\_oneclass\_division}};
    \node[entry] at (3, 5.5) () {\emph{configure\_division\_outlier}};
    
    \node[entry] at (3, 4) () {\emph{get\_train\_test}};
    \node[entry] at (3, 3) () {\emph{get\_train\_test\_outlier}};
    
    \draw[dotted, thick] (1,4.5) -- (10,4.5);
    \draw[dotted, thick] (1,8.5) -- (10,8.5);
	
	\draw [block, fill=white] (6.25,5.25) rectangle (9.75,10.75);

	\node[label] at (6.25,8) () {\bfseries Base datasets};
	
    \node[entry] at (8.25,10) () {\emph{MNIST\_base}};
    \node[entry] at (8.25, 9) () {\emph{Omniglot\_base}};
    \node[entry] at (8.25, 8) () {\emph{CIFAR10\_base}};
    \node[entry] at (8.25, 7) () {\emph{CIFAR100\_base}};
    \node[entry] at (8.25, 6) () {\emph{SVHN\_base}};

	\draw [block, ultra thick, fill=yellow!50] (1,12) rectangle (5,15);

	\node[label] at (1,13.5) () {\bfseries Models};
	
	\node[entry] at (3,14.5) () {\emph{OSRModule}};
    \node[entry] at (3,13.5) () {\emph{TSoftmax}};
    \node[entry] at (3,12.5) () {\emph{Openmax}};	
	
	\draw [block] (6,12) rectangle (10,17);
	
	\node[label] at (6,14.5) () {\bfseries Utils};
	
    \node[entry] at (8,16.5) () {\emph{get\_openmax\_epsilon}};
    \node[entry] at (8,15.5) () {\emph{get\_softmax\_epsilon}};
	\node[entry] at (8,14.5) () {\emph{onehot\_bg}};
    \node[entry] at (8,13.5) () {\emph{inverse\_transform}};
    \node[entry] at (8,12.5) () {\emph{grayscale\_transform}};
	
	\draw [block] (1,16) rectangle (5,19);
	
	\node[label] at (1,17.5) () {\bfseries Architectures};
	
    \node[entry] at (3,18.5) () {\emph{fc\_lower\_stack}};
    \node[entry] at (3,17.5) () {\emph{osrci\_lower\_stack}};
    \node[entry] at (3,16.5) () {\emph{alexNet32\_lower\_stack}};

    \draw[arr] (8,12) -> (8,11.25);
    \draw[arr] (6,13.5) -> (5.25,13.5);

    \draw[arr] (3,16) -> (3,15.25);

    \draw[arr] (3,11) -> (3,11.75);

\end{tikzpicture}}
    \caption{Package architecture in modules}
    \label{fig:schema}
\end{figure}
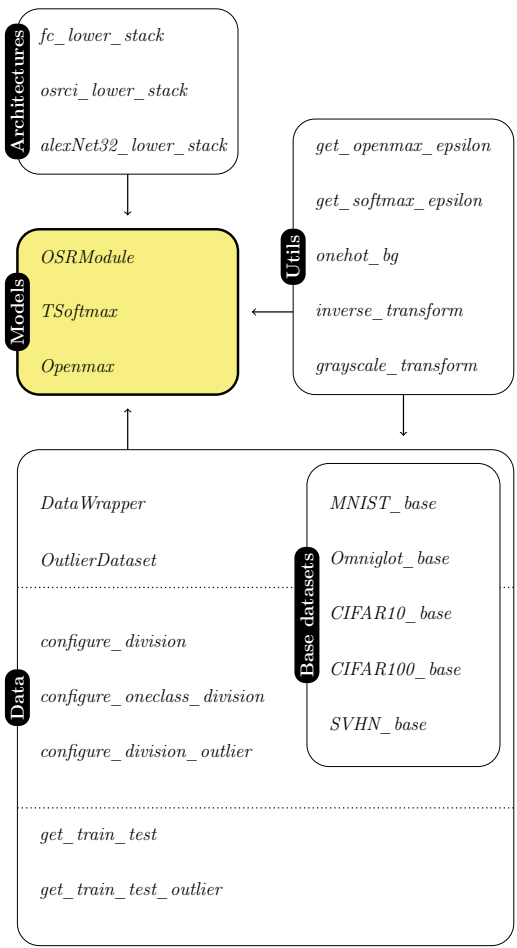

Figure~\ref{fig:schema} shows the package architecture. The central module, marked in yellow, is the \textit{Models} module, which contains the \textit{OSRModule} abstract class and the \textit{Thresholded Softmax} and \textit{Openmax} methods. The architectures module contains three exemplary \textit{lower-stack} architectures, being a primary and necessary part of the final neural network architecture used in OSR. 

A significant part of the \textit{torchosr} package is the \textit{Data} module, which allows for downloading and managing datasets. The sub-module \textit{Base Datasets} contains five classes to handle datasets. The base class implementation was taken from the \textit{PyTorch} library. Classes have been modified to enable processing all data without obligatory division into test and training subsets. Methods have been added to unify the handling of datasets. In addition, the \textit{Data} module provides two classes that allow the use of base sets in experimental OSR protocols -- \textit{DataWrapper} and \textit{OutlierDataset}. The module contains functions that allow generating a given number of configurations of known and unknown classes based on a base dataset and creating consecutive derived sets divided into folds. The \textit{Utils} module contains tools useful for specifying method hyperparameters and transforming datasets.

\subsection{Modules}

The package is divided into five modules. Two of them (\textit{Base Datasets} and \textit{Data}) are responsible for handling data sets and preparing configurations for experimental analysis of methods. The following two modules (\textit{Models} and \textit{Architectures}) contain methods and their components capable of recognizing in Open Space. The \textit{Utils} module includes additional valuable tools for handling datasets and specifying method hyperparameters.

\subsubsection{Base Datasets Module}

The \textit{Base Datasets} module contains modified implementations of datasets from the \textit{PyTorch} library. The training and test sets were homogenized in all employed datasets to have all available instances for the cross-validation procedure.

The module contains the following datasets:
\begin{itemize}
     \item \textit{MNIST\_base} -- An MNIST dataset representing handwritten digits. The collection consists of 10 classes, each containing 7,000 objects. With modern neural network architectures and computational capabilities, the set allows achieving nearly 100\% quality of the classed-set classification. However, when considering some classes as unknown, the dataset states a research challenge.
     \item \textit{Omniglot\_base} -- The collection represents handwritten letters. It consists of 964 classes, each described by 20 objects. The set is often used in Open Set Recognition as an Outlier dataset -- its instances are used as objects of the unknown class.
     \item \textit{CIFAR10\_base} -- A collection of low-resolution color images of objects from 10 classes, 6,000 objects from each class.
     \item \textit{CIFAR100\_bas}e -- CIFAR10 extension to 100 classes, 600 objects each.
     \item \textit{SVHN\_base} -- A collection of nearly 100,000 objects from 10 classes, illustrating text fragments depicting numbers.
\end{itemize}

\subsubsection{Data Module}

The data module contains methods useful for method evaluation in Open Space. It contains two dataset extension classes with additional utilities, functions for specifying class configurations for experimental analysis, and functions for cross-validation splits.

The module includes the following extension classes:
\begin{itemize}
     \item \textit{DataWrapper} -- A class that extends base datasets to allow utilizing them in OSR tasks. It has parameters to specify which classes are considered KKC and UUC and to select a specific subset of the instances. When specifying known classes, if the identifiers of these classes are not consecutive integers, the classes will be re-indexed so that KKC are described by the indices $[0, k-1]$, where $k$ is the number of known classes. All samples of unknown classes will get a common identifier $k$. The class can handle one-hot encoding of class labels.
     \item \textit{OutlierDataset} -- A class that allows specifying a set containing instances of two base datasets, in which the classes from one set are considered as KKC and from the second set as UUC. The class has parameters allowing data shuffle, on-hot encoding, and defining labels for unknown classes.
\end{itemize}

and the following functions:
\begin{itemize}
     \item \textit{configure\_division} -- A function that allows specifying the desired number of data set configurations in an experimental Holdout protocol. It accepts a \textit{VisionDataset} and the following parameters: \textit{n\_openness} describing the generated number of \textit{Openness} of the problem (depending on the size of the KKC and UUC), \textit{repeats} describing the number of random class assignments to the KKC and UUC categories. When drawing \textit{Openness}, configurations with a larger total number of classes are more likely to be drawn. For example, with \textit{n\_openness} equal to 10 and \textit{repeats} equal to 5 -- 50 configurations will be generated. The method returns a list containing class indexes falling into the KKC and UUC categories and a list describing the Openness of individual configurations.
     \item \textit{configure\_oneclass\_division} -- A modification of the \textit{configure\_division} function, where only one class will belong to KKC.
     \item \textit{configure\_division\_outlier} -- A modification of the \textit{configure\_division} function dedicated to the Outlier experimental protocol. The method randomly selects KKC from \textit{base\_dataset} and UUC from \textit{outlier\_dataset}.
     \item \textit{get\_train\_test} -- A function that specifies a subset of data describing a single cross-validation fold in a Holdout experimental protocol. It accepts as parameters a base \textit{VisionDataset}, the number of folds, the index of the current fold, and KKC and UUC identifiers. The method uses \textit{DataWrapper} to prepare a dataset with a specific configuration. The function flag \textit{tunning} determines whether only 10\% of instances (dedicated to hyperparameter optimization) will be considered. When the flag is set to \textit{False}, the remaining 90\% of the set will be considered. Then the set is divided into folds, and only instances from the current subset specified by the \textit{fold} parameter are selected. UUC samples are removed from the training data.
     \item \textit{get\_train\_test\_outlier} -- A function analogous to \textit{get\_train\_test}, designed for splitting files processed in Outlier protocol type. The method uses the \textit{DataWrapper} class to specify a file containing only KKC, where the labels are one-hot encoded, and the \textit{OutlierDataset} class to combine instances of known and unknown classes in the test set.
\end{itemize}

\begin{figure}[!htb]
    \centering
    \includegraphics[width=0.45\textwidth]{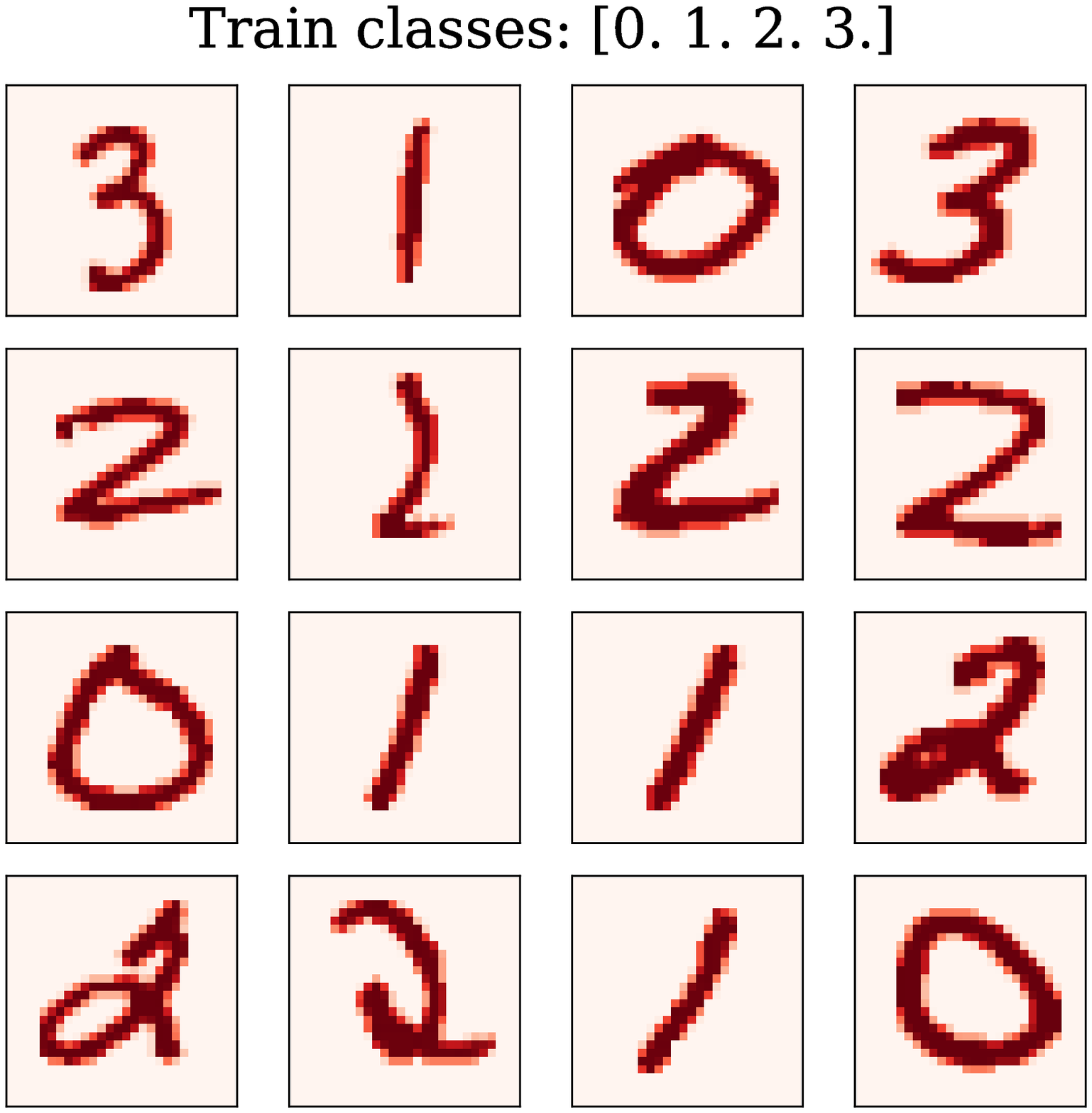}
    \includegraphics[width=0.45\textwidth]{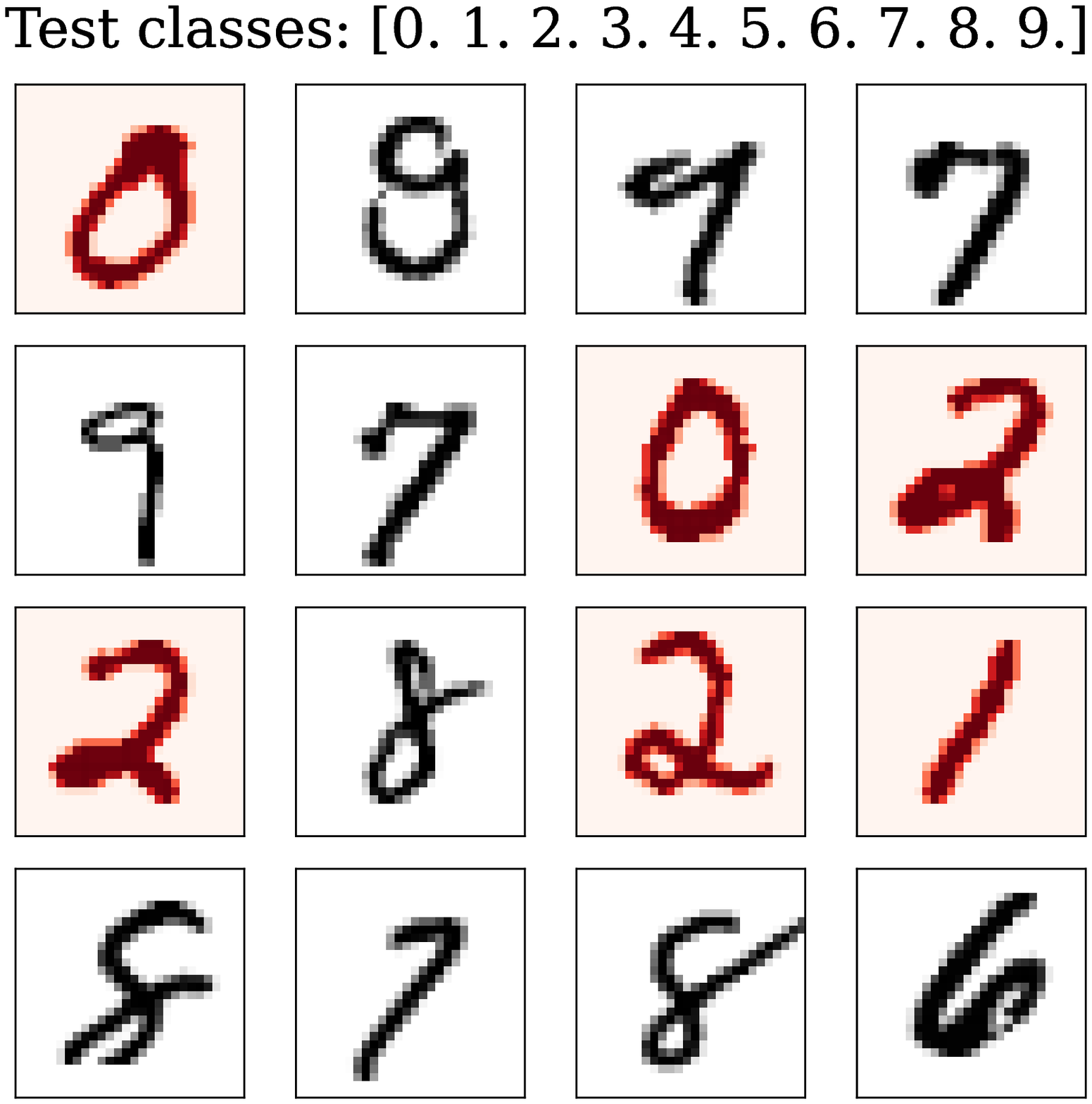}
    \caption{Examples from training and testing sets}
    \label{fig:traintest}
\end{figure}

Figure \ref{fig:traintest} shows nine objects from the training and testing sets. KKC objects are marked in red, and UUC in blue. Only objects of known classes are available in the training set, whereas both known and unknown classes are in the test set.

\subsubsection{Models Module}

The \textit{Models} module contains implementations of two state-of-the-art Open Set Recognition methods. \textit{Openmax} \cite{bendale2016towards} is the first published method dedicated to Open Set Recognition using neural networks. The second method is \textit{Thresholded Softmax} -- a solution considered to be the baseline in the field, the operation of which is based on probability thresholding on the network's last layer. At initialization, methods take on the \textit{lower stack} network structure, which is the primary part of the architecture, extended by \textit{upper stack} depending on the method.

Below is a methods' operation and method-specific parameters description.

\begin{itemize}

    \item Thresholded Softmax (\textit{TSoftmax}) -- Modification of the standard neural network with Softmax final layer, in which objects with dominant supports below the \textit{epsilon} threshold will be recognized as unknown objects. At initialization, it accepts a \textit{lower stack} structure, the number of known classes, and a threshold parameter \textit{epsilon}. \textit{Upper stack} methods consist of one fully connected layer. Only KKC objects are used during the training procedure, and the final architecture is a combination of \textit{lower stack} and \textit{upper stack}. Additionally, the normalization of activation values at the output from the structure was applied \cite{neal2018open}.
     In the testing procedure, the output values are calculated for all objects (KKC and UUC in the test set). Then the Softmax activation function is applied, which computes prediction probabilities. Within the KKC, the object will be recognized as the one with the most significant support, whereas objects will be recognized as UUC when the leading probability for the KKC is less than \textit{epsilon}. \textit{Thresholded Softmax} will recognize \textit{uncertain} objects -- those that are close to the decision boundary -- as unknown. It is worth noting that this method does not fully solve Open Set Recognition task, as objects located far from the decision boundary will still be recognized as KKC, which is related to \textit{Unbounded Open Space} \cite{boult2019learning}. Nevertheless, this method is the baseline for most experimental evaluations of OSR methods.
     
    \item \textit{Openmax} -- The first solution dedicated to Open Set Recognition using Neural Networks. It is an alternative to the standard Softmax layer, limiting the KKC recognition space and guaranteeing \textit{Bounded Open Space}. During initialization, it accepts \textit{lower stack}, the number of known classes, the \textit{epsilon} threshold, and the parameters \textit{tail} and \textit{alpha}, specifying respectively the tail size for estimating the Weibull distribution and the number of most significant classes taken into account. During the training of the network, apart from the standard optimization of the weights, the \textit{fitting} procedure of Weibull distributions to the KKC is carried out. Mean Actiavation Vectors (MAV) are calculated for the correctly recognized samples, then their distance to the MAV is calculated to determine the distance distribution. The Euclidean distance is calculated. The distances are sorted, and the \textit{tail} of the most distant observations is selected. The Weibull distribution is fitted to the \textit{tail} distances.
     In the testing procedure, the activation at the network output and the sum of activation (for all KKC) for individual samples are calculated. On the basis of Weibull distributions fitted in the training procedure for \textit{alpha} most significant classes, \textit{pseudoactivations} for UUC are calculated. Then, as in \textit{TSoftmax}, the activations are subjected to the Softmax function, and probabilities and predictions are determined. \textit{Openmax} additionally uses the mechanism described in \textit{TSoftmax} -- the method makes predictions in favor of UUC when the pseudo-activation for UUC is the highest or when the maximum support for KKC is less than the \textit{epsilon} parameter. A more detailed description of the method can be found in the publication introducing the solution \cite{bendale2016towards}.
     
\end{itemize}

Four Weighted Accuracy metrics, built into the models, are used during the testing procedure. The metrics were designed to describe as accurately as possible the potential of the models to combine closed-set classification tasks and to recognize unknown objects. In addition, during the testing procedure, both models have the \textit{conf} parameter that allows returning, in addition to the metric values, the confusion matrix for the given test data. Based on these confusion matrices, it will be possible to calculate a larger set of metrics.

\begin{figure}[!htb]
    \centering
    \includegraphics[width=\textwidth]{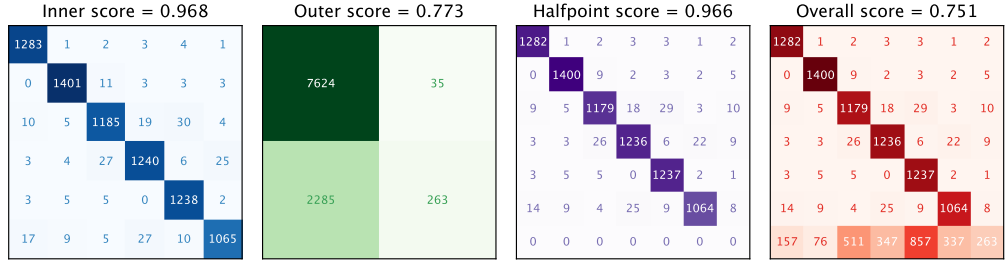}
    \caption{Sample confusion matrices and the results of individual metrics obtained on their basis}
    \label{fig:matrixes}
\end{figure}

The following metrics are calculated in the testing procedure:
\begin{itemize}
     \item \textit{Inner score} --  Measures the closed-set classification performance. Predictions are made only within known class labels, and when calculating the metric, UUC test objects are not involved in determining its value.
     \item \textit{Outer score} -- The ability to recognize objects of unknown classes, referred to as the binary ability to distinguish between KKC and UUC. Known classes are marked as positive and unknown as negative \cite{scheirer2012toward} samples.
     \item \textit{Halfpoint score} -- Modification of the Inner score, considering False Unknowns -- objects that belong to the KKC but were incorrectly recognized as UUC.
     \item \textit{Overall score} -- Measures the ability to classify known and unknown objects. UUC in the testing process will be equivalent to the additional KKC class.
\end{itemize}
The first two metrics (usually obtained using F-score or AUC \cite{dhamija2022five}) are used to report the quality of models in most publications. The presented \textit{torchosr} package has extended the standard solutions with two metrics that give a broader insight into how the models work. Figure \ref{fig:matrixes} shows example confusion matrices and the metrics calculated for them.

\subsubsection{Architectures Module}

The Architectures module contains three exemplary architectures used as the \textit{lower stack}. This part of the network architecture, of the \textit{PyTorch Sequential} type, is the primary and first stage of processing, supplemented by \textit{Openmax} and \textit{Thresholded Softmax} method-specific end layers. The inclusion of baseline architectures in the package is intended to facilitate the comparison process, including methods to be published in the future based on evaluation with \textit{torchosr}. All methods that return architectures accept the \textit{n\_out\_channels} parameter, specifying the output size to which the input data will be transformed.

The module includes the following architectures:
\begin{itemize}
     \item \textit{fc\_lower\_stack} -- The function additionally accepts parameters describing the color depth and the size of the image. This architecture converts the image data into a vector at the beginning of processing. The returned network comprises one fully connected layer and the ReLU activation function.
     \item \textit{osrci\_lower\_stack} -- This function also accepts the image's size and color depth parameters. The returned architecture is inspired by the simple network setup often used to evaluate OSR methods \cite{neal2018open}. The architecture includes, two convolutional layers, a dropout, and a max pooling layer.
     \item \textit{alexNet32\_lower\_stack} -- function returns network inspired by the \textit{AlexNet} \cite{krizhevsky2017imagenet} architecture, simplified to handle smaller (32 x 32 pixel) color images. It includes five convolutions layers.
\end{itemize}

\subsubsection{Utils Module}

The \textit{Utils} module is intended for additional functionalities. In the current version of the software, its functions are related to the preprocessing of datasets and the optimization of method hyperparameters. The module contains the following components:
\begin{itemize}
     \item \textit{onehot\_bg} -- A function for transforming labels into a one-hot encoded version.
     \item \textit{inverse\_transform} -- A function that inverts the image intensity for images where a single pixel is represented by a floating-point value in the 0-1 range.
     \item \textit{grayscale\_transform} -- A function for transforming color images to grayscale, obtained by averaging the image's color channels.
     \item \textit{get\_softmax\_epsilon} -- A function that returns the suboptimal value of the epsilon parameter for the \textit{Thresholded Softmax} method depending on the number of known classes.
     \item \textit{get\_openmax\_epsilon} -- An analogous method, returning a suboptimal parameter value for the \textit{Openmax} method. Suboptimal values were determined based on a preliminary review of hyperparameters for 180 configurations of the MNIST dataset.
\end{itemize}

\subsection{Processing example}

\emph{Torchosr} module is open package released under the \emph{GPL-3.0} license and versioned in the public \emph{Python Package Index} (PyPI) repository. Therefore, it can be obtained with the \emph{pip} package installer with the command:

\begin{lstlisting}[language=bash,numbers=none]
> pip install torchosr
\end{lstlisting}

It is also possible to download and install the library from the sources, if necessary, to modify the functionality offered by the built-in models and methods.

\begin{lstlisting}[language=bash,numbers=none]
> git clone https://github.com/w4k2/torchosr
> cd torchosr
> make install
\end{lstlisting}

The module can be imported in the standard Python manner.

\begin{lstlisting}[language=Python]
# Importing torchosr
import torchosr
\end{lstlisting}

The code below allows loading the \verb|MNIST_base| dataset.

\begin{lstlisting}[language=Python]
# Import transforms for pre-processing
from torchvision import transforms

# Load MNIST dataset
data = torchosr.data.base_datasets.MNIST_base(root='data', download=True, transform=transforms.Compose([transforms.Resize(28),transforms.ToTensor()]))
>       Dataset MNIST_base
>       Number of datapoints: 70000
>       Root location: data
\end{lstlisting}

Then, for the loaded file, the \textit{configure\_division} function will generate configurations for derived OSR datasets. The sample code generates nine configurations -- three class assignments for three \textit{Openness} each.

\begin{lstlisting}[language=Python]
# Generate OSR problem configurations
config, openness = torchosr.data.configure_division(data,
                                                    n_openness=3, 
                                                    repeats=3, 
                                                    seed=1234)
# Print configurations
for i, (kkc, uuc) in enumerate(config):
    print('C%i - Op: %.3f KKC:%s \t UUC:%s' % (
        i, 
        openness[int(i/3)].detach().numpy(), 
        kkc.detach().numpy(), 
        uuc.detach().numpy()))

> C0 - Op: 0.047 KKC:[0 1 7 9 3] 	 UUC:[6]
> C1 - Op: 0.047 KKC:[6 4 9 2 7] 	 UUC:[1]
> C2 - Op: 0.047 KKC:[1 6 7 0 5] 	 UUC:[9]
> C3 - Op: 0.225 KKC:[8 4 5] 	     UUC:[3 1 9 6]
> C4 - Op: 0.225 KKC:[9 7 4] 	     UUC:[2 5 0 8]
> C5 - Op: 0.225 KKC:[0 4 2] 	     UUC:[3 9 6 8]
> C6 - Op: 0.397 KKC:[3 6] 	         UUC:[9 1 4 0 8 7 5]
> C7 - Op: 0.397 KKC:[2 5] 	         UUC:[9 8 6 3 1 4 7]
> C8 - Op: 0.397 KKC:[4 1] 	         UUC:[3 0 5 9 2 7 8]
\end{lstlisting}

The next step is determining the actual training and test set for the evaluation. The \textit{get\_train\_test} method will be used for this from data modules. In the example code, the division was made for the first of the nine generated configurations and the first of the five folds.

\begin{lstlisting}[language=Python]
# Import DataLoader
from torch.utils.data import DataLoader

# Select KKC and UUC from configuration
kkc, uuc = config[0]

# Get training and testing data for first out of 5 folds
train_data, test_data = torchosr.data.get_train_test(data, kkc, uuc, root = 'data', tunning = False, fold = 0, n_folds = 5, seed = 1234)

# Create DataLoaders
train_data_loader = DataLoader(train_data, batch_size=64, shuffle=True)
test_data_loader = DataLoader(test_data, batch_size=64, shuffle=True)
\end{lstlisting}

For the purpose of presentation, labels of objects located in the training and test data loaders were displayed. By default, labels are transformed using the one-hot encoder. In the test subset, the last label represents objects of an unknown class. The classes have been re-indexed in both subsets so that their labels are consecutive integers.

\begin{lstlisting}[language=Python]
import numpy as np

# Load first batch of Train data and print unique labels
X, y = next(iter(train_data_loader))
print('Train labels:', np.unique(np.argmax(y, axis=1)))

# Load first batch of Test data and print unique labels
X, y = next(iter(test_data_loader))
print('Test labels:', np.unique(np.argmax(y, axis=1)))

> Train labels: [0 1 2 3 4]
> Test labels: [0 1 2 3 4 5]
\end{lstlisting}

The method of initializing the \textit{TSoftmax} method is presented below. The simplest architecture available in the package (consisting only of fully connected layers) was used. The \textit{depth} and \textit{img\_size\_x} parameters describe the dimensions of the images in the MNIST set. The epsilon parameter was determined using a method available in the \textit{Utils} module, which returns a suboptimal parameter value for a given KKC cardinality.

\begin{lstlisting}[language=Python]
# Initialize lower stack
ls = torchosr.architectures.fc_lower_stack(depth=1, img_size_x=28, n_out_channels=64)

# Get epsilon parameter for given number of KKC
epsilon = torchosr.utils.base.get_softmax_epsilon(len(kkc))

# Initialize method
method = torchosr.models.TSoftmax(lower_stack=ls, n_known=len(kkc), epsilon=epsilon)
\end{lstlisting}

It is possible to further proceed with evaluation of the model for the given data. In the example, the number of epochs and the learning rate were defined, a table for the results from subsequent epochs was created, and the loss function and optimizer were defined. In a loop, for each epoch, the training and testing procedure was carried out. The values returned by the test method (\textit{Inner}, \textit{Outer}, \textit{Halfpoint} and \textit{Overall} scores, respectively) were saved to the table.

\begin{lstlisting}[language=Python]
import torch

# Specify processing parameters
epochs = 128
learning_rate = 1e-3

# Prepare array for results
results = torch.zeros((4,epochs))

# Initialize loss function
loss_fn = torch.nn.CrossEntropyLoss()

# Initialize optimizer
optimizer = torch.optim.SGD(method.parameters(), lr=learning_rate)

for t in range(epochs):
    # Train
    method.train(train_data_loader, loss_fn, optimizer)
    
    # Test
    inner_score, outer_score, hp_score, overall_score = method.test(test_data_loader, loss_fn)
    results[:, t] = torch.tensor([inner_score, outer_score, hp_score, overall_score])
      
\end{lstlisting}

\begin{figure}[!htb]
    \centering
    \includegraphics[width=\textwidth]{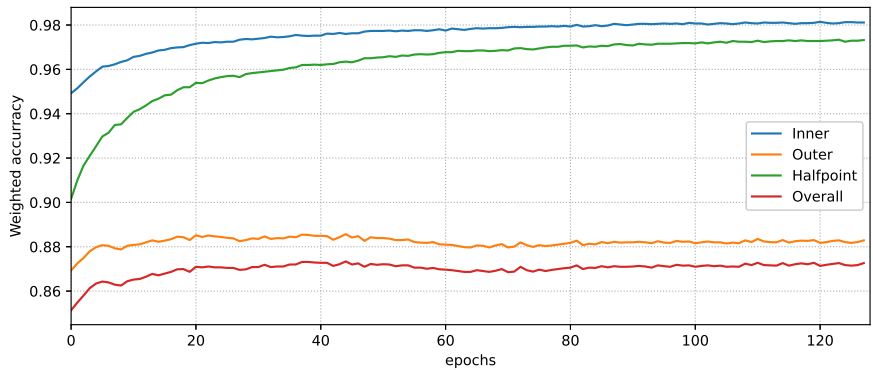}
    \caption{Results of \textit{TSoftmax} exemplary processing}
    \label{fig:example}
\end{figure}

The results of the single processing can be visualized using \textit{matplotlib} library. The output of code presented below is shown in Figure \ref{fig:example}.

\begin{lstlisting}[language=Python]
import matplotlib.pyplot as plt

# Present results
fig, ax = plt.subplots(1,1,figsize=(10,4))
ax.plot(results.T, label=['Inner', 'Outer', 'Halfpoint', 'Overall'])
ax.legend()
ax.grid(ls=':')
ax.set_xlabel('epochs')
ax.set_ylabel('Weighted accurracy')
ax.set_xlim(0,epochs)
\end{lstlisting}

During the test procedure, one can also request a confusion matrix by using the \textit{conf} flag in the test routine.

\begin{lstlisting}[language=Python]
# Call of test method with conf flag
inner_score, outer_score, hp_score, overall_score, \
    inner_c, outer_c, hp_c, overall_c = method.test(test_data_loader, loss_fn, conf=True)

# Print overall confusion matrix
print(overall_c.detach().numpy())
>  [[1244,    2,    1,    1,    3,   12],
    [   1, 1406,    6,    2,    1,   12],
    [   1,    2, 1240,    6,    3,   25],
    [   0,    5,    5, 1303,    7,   25],
    [   5,    4,   18,    7, 1206,   22],
    [ 367,  111,   76,   14,  250,  411]]
\end{lstlisting}

It should be emphasized that the example code presented here is only a fragment of a proper experimental analysis, which should consist of the analysis of many configurations and many iterations of evaluation using cross-validation.

\section{Comparison with available modules}

The Open Set Recognition problem is such a specific field of data processing that its models do not find default implementations in standard libraries dedicated to the issue of pattern recognition, such as \textit{scikit-learn} or \textit{PyTorch}. They contain only certain related models, which are more in the field of outlier and out-of-distribution detection.

Most publications in the field, as long as they contain sources that allow replication of research, do not provide integral estimators or research modules, but rather small portions of the code transforming the outputs of standard solutions. Some original publications contain implementations of methods \footnote{\url{https://github. com/abhijitbendale/OSDN}} \footnote{\url{https://github.com/lwneal/counterfactual-open-set}}, but they are specific solutions for computing packages that are no longer used.

Therefore, for any state-of-the-art surveys not limited to datasets and their divisions available elsewhere, it will be necessary to use some open-source solution that is less popular than the mainstream ML packages.

There are only 38 repositories on GitHub with an open-set-recognition topic. The most popular of these, with just over 800 stargazers\footnote{\url{https://github.com/iCGY96/awesome_OpenSetRecognition_list}}, is a collection of publications and additional materials related to the Open Set Recognition field. Among the available repositories containing proper method implementations, the Generalized Out-of-Distribution Detection Framework \cite{yang2022openood} seems to be the most interesting, containing over 30 algorithms dedicated to the broad phenomenon of Out of distribution detection.

Unfortunately, the wide application of this package means that the subject of Open Set Recognition is only a side element. The package contains an implementation of three methods and does not consider the phenomenon's specificity in the context of a reliable experimental protocol. The ten benchmark sets available within it are limited to predetermined single divisions, effectively hindering a proper review of the Openness of the problem.

In the resources of the GitHub website, at least two small packages are dedicated directly to the OSR phenomenon. In one of them\footnote{\url{https://github.com/ma-xu/Open-Set-Recognition}} -- unfinished and abandoned two years ago -- presents an implementation of the \textit{Openmax} method and several benchmark models. However, unfortunately, the repository is not a coherent, documented package with a fixed structure. The context of the second one is similar\footnote{\url{https://github.com/taslimisina/osr-ood-ad-methods}}, which has a more structured package design but was abandoned a year ago without ensuring the implementation of any basic methods from the field.

Thus, several projects on the open source market address OSR problems, but they are either a side area of the main focus of the package (\textit{OpenOOD}) or have already been abandoned by the developers before being brought to a version that allows for reliable peer-review experiments.

\section{Impact}

In recent years, deep neural networks have already demonstrated recognition capabilities exceeding human perception \cite{he2015delving}. This event has correlated with the growing popularity and trust in \textit{Deep Learning} solutions. The growing trust in these solutions and their limited explainability \cite{marcinkevivcs2020interpretability} make the capability of models to recognize in Open Space particularly important. Due to the variability of concepts and problem definitions, where new classes may appear \cite{mittal2021essentials}, models should be ready to react to unknown objects. Preparing a model for all entries from the \textit{Open World} is impossible.

Adversarial attacks \cite{szegedy2013intriguing}, in which human-imperceptible modifications of input data fool deep models, resulting in incorrect recognition with high support, are a significant threat related to unsupervised extraction of features by deep networks. OSR algorithms have already demonstrated the ability to detect manipulated data and mark them as unknown objects \cite{bendale2016towards}.

Due to the computational costs associated with training deep neural networks, approaches based on \textit{Transfer Learning} \cite{perera2019deep} are also used, where the trained neural network can be used  in multiple domains. With the unsupervised mapping of input features to the deep space \cite{boult2019learning}, objects from unknown classes are represented in the same space as known classes for which the recognition model was constructed. This further emphasizes the importance of evaluating methods in \textit{Open Space}.

The issue of Open Set Recognition is dealt with by a broad group of researchers, which is confirmed by the emergence of a large group of methods in the field and the survey works in recent years \cite{geng2020recent,salehi2021unified,mahdavi2021survey}. Due to the importance of the field and the observed interest from the research community, we hope that the presented package will find its application in many future studies.

\section{Conclusions}

The publication presents the \textit{torchosr} package, compatible with the \textit{PyTorch} library, containing sets, tools, and models useful in the evaluation of the Open Set Recognition task.

The package includes, among others, (1) five classes that handle basic datasets, (2) two dataset extension models, adapting the datasets to the evaluation of OSR tasks in the Holdout and Outlier experimental protocol, (3) methods for determining a set of dataset configurations, and dividing them in cross-validation, (4) two state-of-the-art methods in the field.

Developing methods in the OSR field is critical in the face of deep neural networks gaining popularity and society's trust. In \textit{Open World}, recognition systems, in addition to effective closed-set classification, should be able to determine that input is unknown -- not belonging to classes used for model training. In addition to proposing models of this type, correct and extensive experimental analysis is crucial.

The \textit{torchosr} package is primarily intended to facilitate and disseminate the correct experimental evaluation of methods in the field of Open Set Recognition.

The classes included in the module, adapting data sets for evaluation in \textit{Open Space}, will facilitate the generation and handling of base sets divided into configurations of KKC and UUC classes. The methods included in the module make it possible to avoid storing derivative sets on a computer disk, which increases the availability of extensive experimental analysis, considering many sets with varying Openness and many assignments of classes to categories.

The open-source implementation of the two most commonly used state-of-the-art methods has a chance to allow for easier comparison with them and, hopefully, will be a source of correct implementation, potentially modified and developed by the machine learning community.

\section*{Acknowledgements}
\noindent
This work was supported by the statutory funds of the Department of Systems and Computer Networks, Faculty of Information and Communication Technology, Wroclaw University of Science and Technology.

\bibliographystyle{elsarticle-num} 
\bibliography{bibliography}

\clearpage
\section*{Required Metadata}
\label{}

\section*{Current executable software version}
\label{}

\begin{table}[!ht]
\caption{Code metadata (mandatory)}
\begin{tabular}{|l|p{6.5cm}|p{6.5cm}|}
\hline
\textbf{Nr.} & \textbf{Code metadata description} & \textbf{Please fill in this column} \\
\hline
C1 & Current code version & 0.1.0 \\
\hline
C2 & Permanent link to code/repository used for this code version & $https://github.com/w4k2/torchosr$ \\
\hline
C3 & Legal Code License & GPL-3.0 \\
\hline
C4 & Code versioning system used & git \\
\hline
C5 & Software code languages, tools, and services used & python \\
\hline
C6 & Compilation requirements, operating environments \& dependencies & \textit{numpy}, \textit{scipy}, \textit{torch}, \textit{torchvision}, \textit{torchmetrics}\\
\hline
C7 & If available Link to developer documentation/manual & https://torchosr.readthedocs.io \\
\hline
C8 & Support email for questions & $joanna.komorniczak@pwr.edu.pl$\\
\hline
\end{tabular}
\end{table}

\end{document}